\pdfoutput=1

\documentclass[11pt]{article}

\usepackage[final]{acl}

\usepackage{times}
\usepackage{latexsym}

\usepackage[T1]{fontenc}

\usepackage[utf8]{inputenc}

\usepackage{microtype}

\usepackage{inconsolata}

\usepackage{graphicx}
\usepackage{xcolor} 

\RequirePackage{graphicx}
\RequirePackage{amssymb,amsmath,bm}
\RequirePackage{textcomp}
\RequirePackage{booktabs}
\RequirePackage[textfont=it,tableposition=top]{caption}
\RequirePackage{siunitx}
\RequirePackage{xspace}
\RequirePackage{url}
\RequirePackage{lipsum}  
\RequirePackage{tipa}
\RequirePackage{enumitem}
\RequirePackage{xkeyval}
\RequirePackage{calc}

\usepackage{acl,amsmath,graphicx}
\usepackage{xcolor}
\usepackage[position=top]{subfig}

\usepackage{tipa}
\usepackage{cite}
\PassOptionsToPackage{hyphens}{url}
\usepackage{hyperref}
\usepackage{cleveref}
\usepackage{multirow}
\usepackage{breqn}

\title{Leveraging Allophony in Self-Supervised Speech Models \\ for Atypical Pronunciation Assessment}

\author{Kwanghee Choi, Eunjung Yeo, Kalvin Chang, Shinji Watanabe, David Mortensen \\ Carnegie Mellon University, USA \\ \texttt{\{kwanghec,swatanab,dmortens\}@andrew.cmu.edu}}

\begin{document}
\maketitle

\begin{abstract} 
Allophony refers to the variation in the phonetic realization of a phoneme based on its phonetic environment. 
Modeling allophones is crucial for atypical pronunciation assessment, which involves distinguishing atypical from typical pronunciations.
However, recent phoneme classifier-based approaches often simplify this by treating various realizations as a single phoneme, bypassing the complexity of modeling allophonic variation. 
Motivated by the acoustic modeling capabilities of frozen self-supervised speech model (S3M) features, we propose MixGoP, a novel approach that leverages Gaussian mixture models to model phoneme distributions with multiple subclusters. 
Our experiments show that MixGoP achieves state-of-the-art performance across four out of five datasets, including dysarthric and non-native speech. 
Our analysis further suggests that S3M features capture allophonic variation more effectively than MFCCs and Mel spectrograms, highlighting the benefits of integrating MixGoP with S3M features.\footnote{The full codebase is available at \url{https://github.com/juice500ml/acoustic-units-for-ood}}
\end{abstract}

\section{Introduction}\label{sec:intro}
A phoneme can be phonetically realized differently depending on its environment, a phenomenon known as \textit{allophony} in phonology \citep{twaddell1952phonemes, ladefoged1965nature, collins2019phoneme}. 
For instance, the English phoneme /\textipa{t}/ exhibits various allophonic realizations: [\textipa{t\textsuperscript{h}}] (aspirated stop) in \textit{tap}, [\textipa{t}] (unaspirated stop) in \textit{stop}, [\textipa{\textfishhookr}] (flap) in \textit{butter}, and [\textipa{P}] (glottal stop) in \textit{kitten}.
Accurately capturing these variations is crucial, as it reflects the full spectrum of phonetic realizations within a phoneme.
It is particularly important for atypical pronunciation assessment \citep{twaddell1952phonemes, jokisch2009multilingual, vidal2019epadb}, as it has to distinguish atypical (out-of-distribution; OOD) from atypical (in-distribution) pronunciations \citep{yeo23_interspeech}.

Before the era of deep neural networks (DNNs), allophones were modeled for speech recognition \citep{sagayama1989phoneme,lee1990allophone,young1994tree}.
However, DNN-based approaches \citep{hu2015improved, yeo23_interspeech} depend on phoneme classifiers that treat speech segments from a single phoneme as a single cluster, avoiding the complexity of modeling allophones.
This is partly due to DNN's strong classification capabilities, which rely on trained hidden features to model individual phonemes well.

\begin{figure}[t]
    \centering
    \includegraphics[width=1.0\columnwidth]{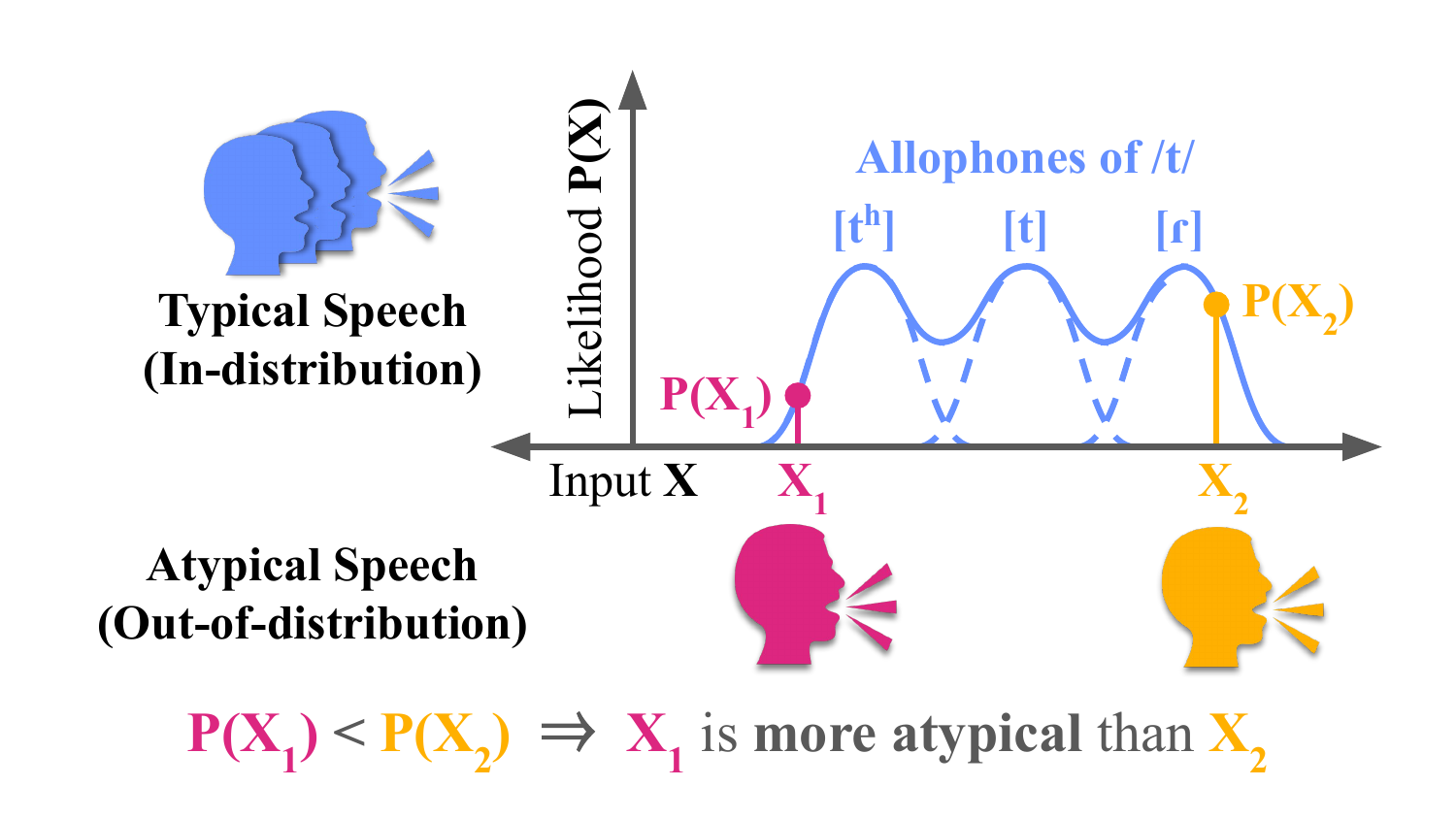}
    \caption{
    Summary of our method, MixGoP.
    We model the likelihood of each phoneme using a Gaussian mixture, trained on typical speech (in-distribution), to capture allophonic variations. 
    We then evaluate on atypical speech (out-of-distribution). 
    The y-axis represents the log-likelihood of a phoneme, where lower values indicate greater atypicality.
    }
    \label{fig:summary}
\end{figure}

In recent years, self-supervised speech models (S3Ms) have shifted the landscape of acoustic modeling.
Unlike DNNs, S3Ms leverage their frozen features directly, without requiring additional training \citep{feng2023superb,chang2024exploring}.
Their effectiveness motivates us to revisit modeling allophones via Gaussian Mixture Models (GMMs) \citep{bilmes1998gentle,young1994tree}.
Consequently, we propose MixGoP, a GMM-based approach that models each phoneme as a set of allophonic subclusters (see \Cref{fig:summary}). 
By integrating GMMs with S3M features, we aim to directly capture the allophonic variations. 
We evaluate MixGoP with S3Ms in atypical pronunciation assessment with dysarthric and non-native speech.

Furthermore, we analyze the S3M features on how well they capture allophonic variation compared to Mel-frequency cepstral coefficients (MFCCs) and Mel spectrograms.
While previous work has shown that S3Ms encode phonetic \citep{pasad2021layer, wells22_interspeech, abdullah2023information, choi2024self} and phonemic information \citep{martin23_interspeech, choi2024understanding}, a detailed investigation of how they capture allophony remains underexplored.

In summary, the contributions of our study are:
\begin{itemize}
    \item MixGoP, a novel pronunciation-assessment approach that considers allophonic variation.
    \item Achieving state-of-the-art performance in four out of five dysarthric and nonnative datasets.
    \item Analysis of the utility of S3M features on MixGoP for capturing allophonic variations.
\end{itemize}

\section{Method}\label{sec:method}
In this section, we briefly review the conventional approach to pronunciation assessment, Goodness of Pronunciation (GoP) \citep{Witt2000PhonelevelPS}.
We highlight its limitations: (i) modeling a phoneme as a single cluster, and (ii) assuming atypical speech are in-distribution with respect to typical speech. 
We then introduce our method, MixGoP, which addresses these limitations by (i) modeling allophonic variation through a mixture distribution and (ii) relaxing in-distribution assumptions by removing the softmax function.

\subsection{What is Goodness of Pronunciation?}\label{sssec:gop}
GoP is a phoneme-level pronunciation score\footnote{While \citet{Witt2000PhonelevelPS} uses the term ``phone''-level pronunciation scores, we use the term ``phoneme'' to emphasize that the unit includes allophones.
Note that \citet{Witt2000PhonelevelPS} also suggests that their use of ``phone'' roughly corresponds to a phoneme.} that measures how much the acoustic output of atypical (dysarthric or nonnative) speech deviates from that of typical speech (healthy or native).
GoP is measured by how likely a speech segment ($\mathbf{s}$) is to be the intended phoneme ($p$).
Given the phoneme classifier $P_\theta(p | \mathbf{s})$ with trainable parameters $\theta$, GoP is measured as the log phoneme posterior,\footnote{GoP is traditionally calculated as the average log probability of phoneme $p$ over the corresponding time frames: $\frac{1}{|F|} \sum_{f \in \mathbf{s}} \log P_\theta (p | f)$, where $f$ refers to frames within the utterance $\mathbf{s}$. In our study, we simplify this by considering the GoP as the log probability of the phoneme as a whole, rather than averaging over frames.}
\begin{align}
    \texttt{GoP}_p(\mathbf{s}) = \log P_\theta(p | \mathbf{s}).
    \label{eq:gop}
\end{align}

\subsection{Limitations of GoP}\label{sssec:limit_gop} 
Conventional phoneme classifiers used in GoP assume a \textit{single cluster} for each phoneme.
This is because logits $f_\theta(\mathbf{s})$ are often modeled with a speech encoder $\texttt{Enc}$ and a subsequent fully-connected (FC) layer with weights $\mathbf{W} \in \mathbb{R}^{|\mathcal{V}| \times F}$ \citep{xu21k_interspeech, yeo23_interspeech}:
\begin{align}
    f_\theta(\mathbf{s}) = \mathbf{W} \cdot \texttt{Enc}(\mathbf{s}) \label{eq:fc}
\end{align}
where $\texttt{Enc}(\mathbf{s}) \in \mathbb{R}^{F}$, $|\mathcal{V}|$ denoting the vocabulary size (total number of phonemes), and $F$ the output dimension of the encoder.
If we consider a frozen encoder, the trainable parameter $\theta = \{\mathbf{W}\}$.
Here, the weights $\mathbf{W}$ can be understood as a codebook, containing a $F$-dim centroid for each phoneme.
It requires a \textit{unimodal} (single peak) clustering of hidden features $\texttt{Enc}(\mathbf{s})$ for each phoneme.
This limits the ability to capture allophonic variation, as allophones are represented as distinct acoustic subclusters within each phoneme.

Another limitation comes from the assumption that observed speech segments are \textit{in-distribution} with respect to the training data. 
This comes from the phoneme classifier $P_\theta$ formulation,
\begin{align}
    P_\theta(p|\mathbf{s}) = \texttt{softmax}(f_\theta(\mathbf{s}))[p]. \label{eq:softmax}
\end{align}
With the phoneme classifier relying on the \texttt{softmax} function, which models a categorical distribution, $\mathbf{s}$ is expected to be within phoneme distribution found in typical speech.
However, this assumption is less suitable for atypical speech, which often exhibits substantial acoustic differences from typical speech \citep{yeo23_interspeech,korzekwa2021mispronunciation}.

\subsection{MixGoP: Modeling multiple subclusters within a single phoneme}\label{sssec:MixGoP}
To address the two limitations presented in \Cref{sssec:limit_gop}, we introduce MixGoP, a mixture distribution-based GoP.

First, to overcome the unimodal assumption, MixGoP replaces phoneme classifier $P_\theta(p | \mathbf{s})$ in \cref{eq:gop} with a Gaussian mixture model (GMM).
GMM is a weighted sum of Gaussian distributions that can directly model the phoneme likelihood $P_\theta (\mathbf{s} | p)$ (distribution of speech segment $\mathbf{s}$ for each individual phoneme $p$).
Accordingly, we formulate the phoneme likelihood as follows:
\begin{align}
    P_\theta (\mathbf{s} | p) = \sum_{c=1}^C \pi^c_{p} \mathcal{N}(\texttt{Enc}(\mathbf{s}) | \pmb{\mu}^{c}_{p}, \mathbf{\Sigma}^{c}_{p} ) \label{eq:GM}
\end{align}
where $\mathcal{N}$ denotes the multivariate Gaussian distribution, $\pmb{\mu}^{c}_{p} \in \mathbb{R}^F$ and $\mathbf{\Sigma}^{c}_{p} \in \mathbb{R}^{F \times F}$ is the mean vector (centroid) and covariance matrix, and $\pi^{c}_{p} \in [0, 1]$ is the mixing coefficient.
Here, the trainable parameter $\theta = \{\pmb{\mu}_p^c, \mathbf{\Sigma}_p^c, \pi_p^c\}_{c\in[C], p\in \mathcal{V}}$.
Then, we can newly define our MixGoP score as:
\begin{align}
    \texttt{MixGoP}_p(s) = \log P_\theta (\mathbf{s} | p).\label{eq:mixgop}
\end{align}
Our MixGoP score differs from the original GoP score in \cref{eq:gop} by replacing the phoneme posterior $P_\theta (p|\mathbf{s})$ with the phoneme likelihood $P_\theta (\mathbf{s} | p)$.
By doing so, we are also removing the influence of phoneme prior $P(p)$, which is known to be effective in practice \citep{yeo23_interspeech}.

Second, MixGoP removes the \texttt{softmax} function of \cref{eq:softmax} by directly using the log-likelihood in \cref{eq:GM,eq:mixgop}.
It relaxes the assumption of phonemes in atypical speech being in-distribution.
The quadratic term inside each Gaussian:
\begin{align}
    -\frac{1}{2} (\texttt{Enc}(\mathbf{s}) - \pmb{\mu}^{c}_{p})^T ({\Sigma^{c}_{p}})^{-1} (\texttt{Enc}(\mathbf{s}) - \pmb{\mu}^{c}_{p}) \label{eq:mahala}
\end{align}
directly relates to the Mahalanobis distance, which is commonly used for OOD detection \citep{lee2018simple}. 
By avoiding the \texttt{softmax}, MixGoP is likely to be more robust in handling OOD speech.

In summary, we train a total of $|\mathcal{V}|$ GMMs (one for each phoneme) where each GMM is composed of $C$ subclusters, \textit{e.g.}, $C=32$.
$C$ is kept constant across all phonemes, as it is known that sufficiently large number of Gaussian mixtures can approximate any probability density \citep{nguyen2020approx}. 
Experiments on the influence of $C$ on downstream performance can be found in \Cref{subsec:abl-clusters}.
We use the k-means algorithm to determine the initial cluster centers and the expectation-maximization (EM) algorithm to optimize the parameters of the Gaussian mixtures, using \texttt{scikit-learn 1.4.1} \citep{scikit-learn}.
By considering allophony in modeling, MixGoP is expected to better reflect the distribution of each phoneme.



\section{Experiments}\label{sec:exp}
\subsection{Datasets}\label{subsec:datasets}
We use five datasets: three dysarthric speech datasets (UASpeech \citep{kim2008dysarthric}, TORGO \citep{rudzicz2012torgo}, and SSNCE \citep{ta2016dysarthric}) and two non-native speech datasets (speechocean762 \citep{speechocean762} and L2-ARCTIC \citep{zhao2018l2}).
In this paper, we use healthy or native speech as the training sets, and dysarthric and non-native speech as the test sets, in line with the OOD literature \citep{DBLP:conf/iclr/HendrycksG17}. 
Refer to \autoref{subsec:dataset-details} for more details.

\subsection{Feature extraction}\label{subsec:features}
For our experiments, we compare various speech feature extractors $\texttt{Enc}(\mathbf{s})$  (\cref{eq:fc,eq:GM}).

\paragraph{Traditional acoustic features.}
We use the Mel-Frequency Cepstral Coefficients (MFCCs) and Mel spectrograms as baselines, using the default hyperparameters of librosa \citep{mcfee2015librosa}.

\paragraph{TDNN-F features.}
We compare with a factorized time-delay neural network (TDNN-F) model \citep{povey2018semi} for the speechocean762 dataset, as TDNN-F features have been often used as baselines \citep{speechocean762,gong2022transformer,chao20223m,do2023hierarchical}.

\paragraph{S3M features.}
We employ two frozen S3Ms: XLS-R-300M \citep{babu22_interspeech} and WavLM-Large \citep{chen2022wavlm}. 
XLS-R (shorthand for XLS-R-300M), trained cross-lingually, has demonstrated strong performance in ASR for low-resource languages \citep{babu22_interspeech} and dysarthric speech assessment \citep{yeo23_interspeech}. 
We also employ WavLM (shorthand for WavLM-Large), a state-of-the-art model for various tasks, including phoneme recognition \citep{feng2023superb,yang21c_interspeech}.

As different layers of S3Ms are known to encode different information \citep{pasad2021layer,pasad2023comparative}, we use features from each layer. 
Specifically, we extract convolutional features (denoted as layer index 0) and all consecutive Transformer features (denoted as layer indices 1 through 24).

\paragraph{Feature segmentation.}
We segment the features according to the start and end timestamps of each phoneme.
Refer to the detailed time-alignment process in \autoref{subsec:dataset-details}.
Then, we apply center pooling to extract one feature per segment.\footnote{We chose center pooling over average pooling to reduce the impact of inaccurate phoneme alignments on GoP calculations.
Both pooling methods are known to encode similar amounts of phonetic information for S3Ms \citep{pasad2023self, choi2024self}.}

\subsection{Baselines}\label{subsec:baseline}
We verify the effectiveness of our MixGoP by comparing it against various baselines \citep{yeo23_interspeech, sun2022out, shahin2023phonological, scholkopf2001estimating}.
We evaluate on all the speech features listed in \Cref{subsec:features} across all the methods for fair comparison.
These baselines are categorized into two groups: (i) phoneme classifier-based and (ii) OOD detector-based approaches.

\paragraph{Phoneme classifier-based approaches} encompass conventional GoP formulations, which assume a unimodal distribution and in-distribution of phonemes, as discussed in \Cref{sssec:limit_gop}. 
We employ four popular GoP formulations\footnote{GoP formulations were named after the underlying models \citep{speechocean762,yeo23_interspeech}, leading to names like GMM-GoP. However, the GMM-GoP scoring method (\cref{eq:gop}) does not necessarily rely on GMM.}: GMM-GoP \citep{Witt2000PhonelevelPS}, NN-GoP \citep{hu2015improved}, DNN-GoP \citep{hu2015improved}, and MaxLogit-GoP \citep{yeo23_interspeech}.
Note that all formulations use the same underlying phoneme classifier $P_\theta(p|\mathbf{s})$.
They only differ by how to calculate the GoP scores.
Refer to \citet{yeo23_interspeech} for more details.

\paragraph{OOD detector-based approaches} calculate GoP by measuring how likely an input is to be an outlier.
In other words, they can quantify the level of atypicalness.
Our MixGoP is one of these approaches, as MixGoP models the likelihood $P_\theta(\mathbf{s}|p)$ with typical speech (\cref{eq:GM}) and identifies outliers (atypical speech) based on their likelihood (\cref{eq:mixgop}). 
We additionally test three baselines: k-nearest neighbors (kNN) \citep{sun2022out}, one-class support vector machine (oSVM) \citep{scholkopf2001estimating}, and phoneme-specific oSVM (p-oSVM) \citep{shahin2019anomaly}. 
While kNN has been utilized for OOD detection \citep{sun2022out}, it has not previously been applied to dysarthric or non-native speech.
Conversely, oSVM and p-oSVM have been applied to the evaluation of both disordered and non-native speech \citep{shahin2019anomaly, shahin2023phonological}.

\subsection{Training details}\label{subsec:training}
\paragraph{Phoneme classifier-based.}
The phoneme classifier in \cref{eq:gop} is trained on features from \Cref{subsec:features} with a single learnable FC layer (\cref{eq:fc}) with the default settings of Adam optimizer \citep{DBLP:journals/corr/KingmaB14} for a maximum of 500 iterations.

\paragraph{OOD detector-based.}
For kNN, we construct a kNN model for each phoneme using the features from \Cref{subsec:features}.
Then, we use the maximum Euclidean distance between the test data feature and the nearest 10\% training data feature as the GoP score, following \citet{sun2022out}.
For oSVM, all phonemes is modeled with a single oSVM model \citep{shahin2023phonological}, while p-oSVM modeled each phoneme as a separate oSVM model.
All oSVM models are trained with features using the default hyperparameters of \texttt{scikit-learn 1.4.1} \citep{scikit-learn}.
Radial basis function was used for both oSVM and p-oSVM.
We use the distance from the hyperplane as the GoP score.

\paragraph{MixGoP.} In our MixGoP framework, we apply random subsampling of 512 features per phoneme.
Empirical analysis indicates that subsampling does not necessarily degrade performance (See \Cref{subsec:ablation}).
The number of subclusters for each phoneme-wise Gaussian mixture is set to 32. 
A detailed investigation into the effect of the number of subclusters on GoP performance is discussed in \Cref{subsec:abl-clusters}.

\subsection{Evaluation}\label{subsec:eval}
As described in \Cref{subsec:features}, we segment the spoken utterance $\mathbf{x}$ phoneme-wise: $\mathbf{x} = \{(p_1, \mathbf{s_1})$, $(p_2, \mathbf{s_2})$, $\cdots$, $(p_N, \mathbf{s_N})\}$, where $p_i$ is the phoneme label, $\mathbf{s}_i$ is the observed speech segment, and $N$ is the total number of phonemes within the utterance.
Following \citet{yeo23_interspeech}, we define the pronunciation score of an utterance $\mathbf{x}$ as:
\begin{align}
    \texttt{Pronunciation}(\mathbf{x}) = \frac{1}{N} \sum_{i=1}^{N} \texttt{GoP}_p(s),\label{eq:1_n}
\end{align}
where the definition of the \texttt{GoP} is different per each method.
That is, the GoP scores are averaged across the utterance.

\begin{table*}[t]
\caption{
Kendall-tau correlation coefficient between the pronunciation scores and the dysfluency/disfluency (absolute value).
Bigger is better.
For S3Ms, the best performance across layers is displayed.
}
\label{tab:main}
\centering
\resizebox{\textwidth}{!}{%
\begin{tabular}{cllcccc|cccc}
\toprule
& \multirow{3}{*}{Dataset} & \multirow{3}{*}{Feature} & \multicolumn{4}{c|}{Phoneme classifier-based} & \multicolumn{4}{c}{Out-of-distribution detector-based} \\
\cmidrule{4-7} \cmidrule{8-11}
& & & GMM- & NN- & DNN- & MaxLogit- & kNN & oSVM & p-oSVM & MixGoP \\
&& & GoP & GoP & GoP & GoP & & & & (Proposed) \\
\midrule
\multirow{12}{*}{\rotatebox[origin=c]{90}{Dysarthric speech}} & \multirow{4}{*}{UASpeech}
& MFCC & 0.428 & 0.410 & 0.361 & 0.430 & 0.418 & 0.107 & 0.105 & 0.182 \\
&& MelSpec & 0.209 & 0.172 & 0.242 & 0.214 & 0.101 & 0.099 & 0.086 & 0.039 \\
&& XLS-R & 0.552 & 0.553 & 0.548 & 0.547 & 0.559 & 0.354 & 0.247 & 0.602 \\
&& WavLM & 0.568 & 0.568 & 0.546 & 0.558 & 0.606 & 0.537 & 0.327 & \textbf{ 0.623 } \\
\cmidrule{2-11}
&\multirow{4}{*}{TORGO}
& MFCC & 0.406 & 0.345 & 0.391 & 0.406 & 0.347 & 0.169 & 0.105 & 0.282 \\
&& MelSpec & 0.271 & 0.262 & 0.150 & 0.271 & 0.287 & 0.211 & 0.241 & 0.196 \\
&& XLS-R & 0.677 & 0.674 & 0.641 & 0.675 & 0.704 & 0.586 & 0.536 & \textbf{ 0.713 } \\
&& WavLM & 0.682 & 0.681 & 0.633 & 0.681 & 0.703 & 0.671 & 0.621 & 0.707 \\
\cmidrule{2-11}
&\multirow{4}{*}{SSNCE}
& MFCC & 0.265 & 0.254 & 0.267 & 0.273 & 0.045 & 0.194 & 0.076 & 0.082 \\
&& MelSpec & 0.183 & 0.161 & 0.051 & 0.187 & 0.154 & 0.114 & 0.106 & 0.174 \\
&& XLS-R & 0.542 & 0.542 & 0.499 & 0.544 & 0.503 & 0.193 & 0.167 & 0.541 \\
&& WavLM & 0.547 & 0.547 & 0.486 & 0.547 & 0.523 & 0.358 & 0.234 & \textbf{ 0.553 } \\
\midrule
\multirow{9}{*}{\rotatebox[origin=c]{90}{Non-native speech}} & \multirow{5}{*}{speechocean762}
& MFCC & 0.390 & 0.375 & 0.255 & 0.405 & 0.322 & 0.202 & 0.111 & 0.126 \\
&& MelSpec & 0.214 & 0.064 & 0.232 & 0.229 & 0.111 & 0.109 & 0.071 & 0.004 \\
&& TDNN-F & 0.400 & 0.356 & 0.243 & 0.360 & 0.361 & 0.099 & 0.001 & 0.197 \\
&& XLS-R & 0.533 & 0.531 & 0.372 & 0.536 & 0.443 & 0.312 & 0.157 & 0.499 \\
&& WavLM & 0.535 & 0.533 & 0.380 & 0.534 & 0.432 & 0.395 & 0.173 & \textbf{ 0.539 } \\
\cmidrule{2-11}
&\multirow{4}{*}{L2-ARCTIC}
& MFCC & 0.136 & 0.141 & 0.119 & 0.119 & 0.042 & 0.004 & 0.034 & 0.043 \\
&& MelSpec & 0.049 & 0.039 & 0.032 & 0.032 & 0.022 & 0.003 & 0.027 & 0.010 \\
&& XLS-R & 0.243 & \textbf{ 0.312 } & 0.191 & 0.191 & 0.168 & 0.037 & 0.067 & 0.152 \\
&& WavLM & 0.240 & 0.269 & 0.196 & 0.196 & 0.189 & 0.082 & 0.078 & 0.182 \\
\bottomrule
\end{tabular}%
}
\end{table*}

Similar to \citet{yeo23_interspeech}, we evaluate performance using the Kendall-tau correlation coefficient between the utterance-level pronunciation scores and the ground truth dysfluency/disfluency scores provided by the dataset. 
While \citet{yeo23_interspeech} used additional training data, our setting only uses the aformentioned datasets.

Unlike other datasets, L2-ARCTIC contains only phoneme-wise mispronunciation detection labels (0 for correct, and 1 for mispronounced).
Therefore, we directly measure the correlation between the predicted phoneme-wise pronunciation scores and the mispronunciation detection labels.

\subsection{Results}\label{subsec:results}
\Cref{tab:main} presents the experimental results, where we report the best-performing layer's results for S3Ms.
Refer to \Cref{subsec:layerwise} for the performance of individual layers.
\Cref{tab:main} shows that our proposed MixGoP method achieved the state-of-the-art performance across all the datasets except L2-ARCTIC.
As for the L2-ARCTIC dataset, the best performance was observed with NN-GoP.

\paragraph{Comparison between features.}
We observe that all other features generally underperform compared to both S3Ms, further highlighting the general effectiveness of frozen S3Ms \citep{yang21c_interspeech}.
It aligns with \citet{choi2022opening}, where S3Ms, unlike MFCCs, stores information as a relative distance between the features so that the Mahalanobis distance of MixGoP (\cref{eq:mahala}) or the Euclidean distance of kNN (\Cref{subsec:gop_vs_knn}) can be effective.
We explore this discussion in detail in \Cref{sec:motiv}.

\paragraph{Comparison between datasets.}
We observe that MixGoP tends to be effective on dysarthric datasets, whereas NN-GoP tends to perform well on non-native datasets.
We suspect this is due to dysarthric test sets being more OOD than non-native test sets, so accurate likelihood estimation becomes more important.
Dysarthric train/test datasets are split by healthy and dysarthric speakers.
However, non-native datasets only contain non-native speakers, where we split train/test using utterance-wise or phoneme-wise pronunciation scores.
(Details in \Cref{subsec:dataset-details}.)
Hence, it is likely that the train/test difference of non-native datasets is less severe than that of dysarthric datasets.
Inherent acoustic differences between dysarthric and non-native speech may have further widen the dataset differences \citep{yeo23_interspeech,korzekwa2021mispronunciation}.

\paragraph{Comparison within the same groups.}
For \textit{phoneme classification-based} baselines, performance greatly differs across methods, despite all four methods being based on the same classifier.
This supports the importance of selecting an appropriate equation for uncertainty quantification in phoneme class-based methods, aligning with the findings of \citet{yeo23_interspeech}.
Regarding \textit{OOD detector-based} baselines, SVM-based methods usually perform the worst across all datasets.
In contrast, kNN achieves performance comparable to our proposed MixGoP on the TORGO and L2-ARCTIC datasets, while MixGoP delivers the best overall performance.

\section{Allophony of S3M features}\label{sec:motiv}
\Cref{sec:exp} empirically demonstrates that leveraging S3M features with MixGoP helps enhance downstream performance compared to other features, such as MFCCs and Mel spectrograms.
This section aims to further verify the suitability of S3Ms for representing individual phonemes with allophonic variations.
First, we examine S3M features at the phoneme-level, using the dimensionality reduction technique in \Cref{subsec:motiv_qual}.
Next, we design a metric to quantify the ability of capturing allophonic variations, comparing S3M features to MFCCs and Mel spectrogram in \Cref{subsec:motiv_quant}.
For our analyses, we used the healthy speech recordings from the TORGO dataset \citep{rudzicz2012torgo}, which includes gold-standard phonemic transcriptions and alignments.

\subsection{Motivating Observation}\label{subsec:motiv_qual}
S3Ms are trained to reconstruct masked signals using surrounding information.
Hence, we hypothesize that this will allow the S3Ms to capture local acoustic characteristics, including allophones from various phonetic environments.
To verify such phenomena, we observed the final layer features of WavLM for each phoneme, which have generally shown the best performance across datasets (See \Cref{fig:layerwise}).
Specifically, we use UMAP dimensionality reduction \citep{mcinnes2018umap} with the cosine distance metric to visualize the features, similar to \citet{choi2022opening}.
We also extract the four utterances closest to each of the ten centroids to observe the phonetic environments of each cluster.

\begin{figure}[t!]
    \centering
    \includegraphics[width=0.44\textwidth]{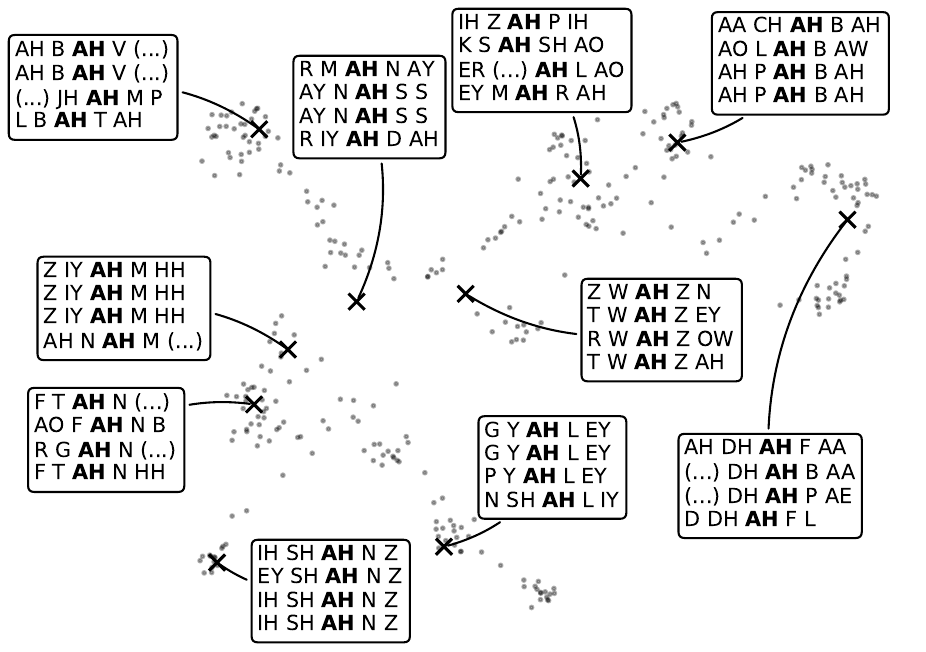}
    \caption{
        Visualization of WavLM-Large features of the \texttt{/AH/} phoneme in the TORGO healthy subset.
        Phonemes are indicated using ARPABET.
        We observe that \texttt{/AH/} consists of subclusters, each reflecting allophones from different surrounding phonetic environments.
    }
    \label{fig:motiv}
\end{figure}

\Cref{fig:motiv} demonstrates one example, with the distribution of \textipa{/2/} (\texttt{/AH/} in ARPABET) and its environments of the healthy subset of TORGO.
We observed multiple clusters for each phoneme, each with phonetically similar environments, which motivates the metric for quantifying allophony ability.

\subsection{Quantifying Allophony}\label{subsec:motiv_quant}
Previous studies have found that S3M features model the phoneme distributions with multiple clusters \citep{wells22_interspeech,martin23_interspeech}.
However, there has been limited analysis on directly quantifying the relationship between the S3M feature subclusters and allophony.
To this end, we design a setting that measures the mutual information between the S3M feature subcluster indices and the surrounding phonetic environment of each phoneme, which is an indicator of allophones.

First, to obtain the subclusters within MFCCs, Mel spectrograms, and S3M features, we apply the k-means algorithm with $k=32$ clusters to the features of each phoneme $V \in \mathcal{V}$.
Then, each utterance has the designated k-means cluster index $I$.
For a dataset with a total of $|\mathcal{V}|$ phonemes, we train $|\mathcal{V}|$ different k-means models.

Note that our MixGoP uses k-means clusters as the initializer.
Also, we observed few to no EM optimization steps due to high dimensionality \citep{wang2015high}.
As a result, the initialized cluster centroids will likely be similar to the final centroids $\pmb{\mu}_p^c$ in \cref{eq:GM} for calculating phoneme likelihood $P_\theta(\mathbf{s}|p)$.

We then compare the utterance-wise cluster indices with their allophony.
Since the TORGO dataset does not provide phonetic transcriptions, we utilize the surrounding phonetic environment, which is closely linked to allophonic variation.
For simplicity, we define the environment $E$ as the natural class of the preceding and following phonemes, similar to phoneme environment clustering \citep{sagayama1989phoneme}.
We use the height, backness, and roundness for the vowels and the place and manner of the consonants for the natural class.
For example, each \textipa{/i/} and \textipa{/k/} is represented as \texttt{close-front-unrounded} and \texttt{velar-plosive}, respectively.
If the phoneme is word-initial or word-final, we also include them as the environment information.
Therefore, with $|\mathcal{C}|$ number of natural classes, the number of all the possible environments is $(|\mathcal{C}| + 1)^2$.

\begin{figure}[t!]
    \centering
    \includegraphics[width=0.3\textwidth]{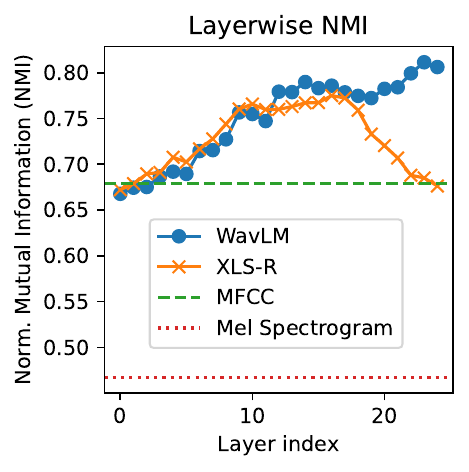}
    \caption{
        Normalized Mutual Information $\text{MI}(I; E)/H(E)$ between the k-means cluster indices $I$ and the phonetic environment $E$ on the TORGO healthy subset.
        We show the layerwise NMI for S3Ms and absolute value for MFCC and Mel spectrogram.
    }
    \label{fig:pnmi}
\end{figure}

To quantify allophonic information within each subclusters, we measure the mutual information $\text{MI}(I; E)$ between the cluster indices $I$ and the environment $E$.
We normalize the value by the environment entropy $H(E)$ so that the resulting value is between 0 and 1.
We call the metric \textit{Allophone environment-Normalized Mutual Information} (ANMI) $\text{MI}(I; E) / H(E)$.
The actual calculation is nearly identical to Phoneme Normalized Mutual Information (PNMI) \citep{hsu2021hubert}, except we replaced phoneme $V$ to environment $E$.

\paragraph{Results.} 
\Cref{fig:pnmi} shows the phonetic environment information inside cluster indices for MFCCs, Mel spectrograms, XLS-R, and WavLM.
For S3M models, we plot across different layers.
We can observe that S3Ms contain more information on the phonetic environment compared to traditional features, implying that S3Ms successfully capture allophony.
This finding introduces an interesting implication regarding the effect of varying cluster sizes when applying k-means to S3M features for discrete units \citep{chang2024exploring}. 
It is known that different cluster sizes lead to varying levels of granularity, with smaller cluster size capturing phoneme information while larger cluster size capturing speaker information \citep{sicherman2023analysing}.
Our results suggest that cluster size in between may capture allophonic variations.

\section{Analysis}\label{sec:discuss}
\subsection{Does capturing phonetic environment leads to better downstream performance?} \label{subsec:pnmi_vs_task}
\begin{figure}[t!]
    \centering
    \includegraphics[width=0.3\textwidth]{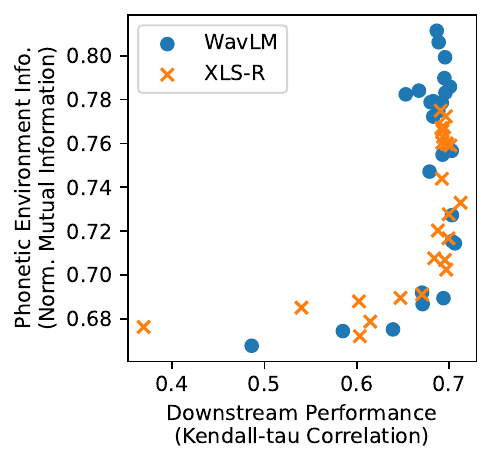}
    \caption{
        Comparing phonetic environment information with the downstream task performance.
    }
    \label{fig:nmi_vs_downstream}
\end{figure}

It is crucial to examine whether capturing phonetic environments (or allophones) actually improves downstream performance. 
To assess this, we compare the amount of phonetic environment information inside S3Ms and the actual downstream performance on the pronunciation assessment.
In \Cref{fig:nmi_vs_downstream}, we observe that the downstream performance positively correlates until around NMI $< 0.72$, where the downstream performance saturates even if the amount of phonetic environment information increases.
We suspect this behavior is due to S3Ms capturing more surrounding information, which may not be useful for the pronunciation assessment task.
Our hypothesis aligns with the previous empirical observation of \citet{pasad2023comparative} and \citet{choi2024self} that S3Ms have non-negligible word-level modeling abilities, which requires a larger temporal receptive field.
Moreover, the layerwise trends of \Cref{fig:pnmi}, \textit{i.e.}, WavLM persistently increasing and XLS-R peaking in the middle, are also similar to previous empirical observations on word-level layerwise information \citep{pasad2023comparative}.

\subsection{Sample efficiency of MixGoP}\label{subsec:ablation}
We randomly subsampled training set samples to check the influence of training data size.
To train the GMM for each phoneme, we can either use all the occurrences in the dataset or limit the maximum number of samples.
For optimal performance, we searched for the maximum number of samples between 64, 128, 256, 512, or using the full dataset.
For example, if we set the maximum as 64, and \textipa{/a/} and \textipa{/i/} each have a total of 100 and 50 samples, to train the GMM of \textipa{/a/}, we randomly subsample 64 samples.
On the other hand, since there are only 50 samples for \textipa{/i/}, all 50 samples are used.

\begin{table}[]
\caption{Ablation on random subsampling.}
\label{tab:numberGM}

\vspace{0.5em}
\resizebox{1.0\linewidth}{!}{%
\begin{tabular}{l|ccccc}
\toprule
Dataset & 64 & 128 & 256 & 512 & Full \\
\midrule
UASpeech & 0.615 & 0.620 & \textbf{ 0.624 } & 0.623 & 0.620 \\
TORGO & 0.704 & 0.709 & 0.712 & \textbf{ 0.713 } & \textbf{ 0.713 } \\
SSNCE & 0.539 & 0.545 & 0.549 & \textbf{ 0.553 } & 0.548 \\
speechocean762 & 0.502 & 0.516 & 0.528 & \textbf{ 0.539 } & 0.536 \\
L2-ARCTIC & 0.182 & 0.178 & 0.181 & 0.182 & \textbf{ 0.197 } \\
\bottomrule
\end{tabular}
}
\end{table}



\Cref{tab:numberGM} shows that increasing the number of training samples generally improves performance. 
However, performance plateaus as the sample size increases, with 512 samples yielding the best results across various datasets, except for L2-ARCTIC. 
This suggests MixGoP performs well even with a relatively small number of samples (fewer than 100), which is advantageous for dysarthric and nonnative speech, where data is often limited.
However, this also indicates that adding more data does not necessarily lead to further improvements, indicating that the model's performance may be constrained by data scalability.

\section{Related works}
\subsection{Phoneme-level Pronunciation Assessment}
\citet{Witt2000PhonelevelPS} first introduced GoP to estimate the log posterior probability of a phoneme using a Hidden Markov Model (HMM). 
Later improvements replaced HMMs with deep neural networks \citep{hu2015improved-slate, hu2015improved, li2016mispronunciation} and S3Ms \citep{xu21k_interspeech, yeo23_interspeech, cao2024framework}. 
GoP has also been enhanced by considering additional factors, such as HMM transition probabilities \citep{sudhakara2019improved, Shi2020ContextawareGO} and phoneme duration \citep{Shi2020ContextawareGO}.

\Cref{sssec:limit_gop} emphasizes the usefulness of framing the pronunciation assessment of atypical speech as the OOD detection task.
\citet{yeo23_interspeech} also addressed this by not using \texttt{softmax} for phoneme classifiers improved performance. 
However, the use of \texttt{softmax} during training introduced in-distribution bias. 
\citet{cheng20_interspeech} modeled input probability with latent representations, but their method still depended on assessment scores to train the prediction model. 
Our approach improves by directly modeling phoneme likelihood instead of relying on phoneme classifiers. 
Furthermore, our approach explicitly accounts for allophonic variation within phonemes.

\subsection{S3M Feature Analysis}
Previous literature on the phonetics and phonology of S3Ms often compared downstream task performance of different layers \citep{martin23_interspeech,pasad2021layer,pasad2023comparative,pasad2023self,choi2024understanding}.
Linear probes \citep{martin23_interspeech,choi2024understanding} or canonical correlation analysis \citep{pasad2021layer,pasad2023comparative,pasad2023self} are often used to measure the amount of information.
Our work is complementary as previous works focus on the existence of the information, whereas we further investigate on how the information is structured within the S3M features.

Also, discrete speech units from k-means clustering of S3M features have been used as the tokenizer for speech \citep{chang2024exploring}.
Its underlying assumption comes from the feature structure being useful, \textit{i.e.}, similar-sounding segments are close to each other.
\citet{choi2024self} showed that phonetically similar words are close to each other.
Also, \citet{baevski2020wav2vec,hsu2021hubert,liu23j_interspeech} demonstrated that phonemes and S3M cluster indices strongly correlate with each other.
\citet{sicherman2023analysing,abdullah2023information} showed that natural classes are also well-clustered.
Finally, \citet{wells22_interspeech} showed that the dynamic nature of a single phoneme articulation is captured by a stream of cluster indices.
Extending previous works, we focus on the multimodal nature of phonemes and demonstrate that allophones are construct subclusters within the single phoneme.

\section{Conclusion}
We demonstrated that improved modeling of allophony can enhance performance in OOD detection for the assessment of atypical speech, and that leveraging S3M features can further improve this performance. 
Specifically, our novel approach, MixGoP, addresses the limitations of uni-modality and in-distribution assumptions by employing Gaussian mixtures, which effectively model allophones and eliminate the need for softmax probabilities. 
Additionally, we show that utilizing S3M features further enhances OOD detection performance. 
Our results also confirm that S3M features capture allophonic variation more effectively than traditional features, validating the extension of our approach to include S3Ms. 
We evaluated eight methods across five dysarthric and nonnative speech datasets, with MixGoP achieving state-of-the-art performance on four of the datasets.

Our work provides a deeper understanding of how S3M representations can be hierarchically structured, from allophones to phonemes.
Further, it sheds new light on the acoustic modeling perspective of speech, expanding the existing k-means-based speech discretization.
It shows the possibility of using atypical speech as a benchmark to measure the quality of S3M features, especially regarding OOD robustness.

\section*{Limitations}
First, a key limitation is the restricted generalizability of our findings across languages. Although we aim for our work to benefit a wide range of atypical speakers, including both dysarthric and non-native speakers, our research primarily focuses on English (four English datasets and one Tamil dataset). This limitation stems from the availability of publicly accessible datasets, but we recognize the need for broader cross-linguistic research in future work to ensure that our findings are applicable across diverse languages.

Additionally, we employed different methods for forced alignment across datasets, as outlined in \Cref{subsec:dataset-details}.
Time alignments were either provided by the dataset or automatically generated using the Montreal Forced Aligner \citep{mcauliffe2017montreal}. 
However, we did not verify the quality of these alignments in our study. 
This introduces the possibility that variations in alignment quality could have impacted the GoP scores, potentially affecting the overall results.
While we do not primarily focus on comparing performance across datasets, future work could benefit from verifying alignment quality to ensure more reliable GoP scores, and cross-dataset comparisons.

We also acknowledge that our allophony analysis was primarily based on the TORGO dataset, which provided time alignments that were manually annotated by linguists. 
Extending this analysis to other datasets with similarly verified time alignments would further support the generalizability of our findings.

Finally, the method used to calculate utterance-level (\Cref{eq:1_n}) pronunciation scores can be improved. 
In our current approach, we simply averaged phoneme probabilities across each utterance; however, it is well-known that certain phonemes have a greater impact on overall pronunciation scores. 
While our initial analysis, as presented in \Cref{subsec:attn},  provides a preliminary exploration of this issue, further investigation is needed to identify more robust approaches.
Expanding upon this analysis could lead to improved techniques that more accurately evaluate atypical speech.

\section*{Ethics Statement}
The risk of atypical pronunciation assessment research primarily pertains to data handling and the potential for unintended consequences in use.

Firstly, while we used publicly available datasets that have undergone prior ethical review, it is important to recognize that these datasets still contain sensitive information, particularly speakers’ voices. 
Since no additional anonymization processes were applied in this study, we strongly recommend that any replication of this work prioritize the protection of participants’ rights and privacy to the greatest extent possible.

Secondly, concerns arise regarding the potential usage of atypical speech assessment scores. 
These assessments may unintentionally reinforce negative stereotypes or stigmas associated with speech disorders or non-native accents. 
If the results are interpreted as evaluations of an individual’s language ability or intelligence, they could further marginalize dysarthric or non-native speakers.
Also, there is a risk in placing too much emphasis on `correctness' in phoneme-level pronunciation assessment. 
Focusing heavily on accurate phoneme production prescribes a rigid, normative standard of speech, potentially penalizing linguistic diversity and variation. 
For both dysarthric and non-native speakers, such an emphasis might overshadow more functional measures of communication success, which may be more meaningful in real-world contexts.
Despite these concerns, which warrant careful consideration, we want to emphasize that our work is intended to have a significant positive impact from an ethical perspective.

Finally, we note that ChatGPT was employed for grammatical refinement and to improve the clarity of English usage in the manuscript.
We also state that every sentence generated by ChatGPT was reviewed by the authors.



\newpage
\section*{Acknowledgments}
This work used the Bridges2 system at PSC and Delta system at NCSA through allocation CIS210014 from the Advanced Cyberinfrastructure Coordination Ecosystem: Services \& Support (ACCESS) program, which is supported by National Science Foundation grants \#2138259, \#2138286, \#2138307, \#2137603, and \#2138296.
\bibliography{custom}

\newpage
\appendix

\section{Datasets}\label{subsec:dataset-details}
In our study, we utilize read speech at the sentence level. 
While the TORGO and SSNCE datasets include both word- and sentence-level data, we focused exclusively on sentences to ensure consistency, as the non-native datasets contain only sentence-level speech. 
Although the UASpeech dataset consists solely of word-level materials, we included it for comparison with prior work that applied GoP to dysarthric speech \citep{yeo23_interspeech}.

In contrast to non-native speech datasets, which provide utterance-level scores, dysarthric speech datasets offer intelligibility scores at the speaker level.
Therefore, we applied these speaker-level scores to each utterance in the dysarthric datasets.

All the datasets are publicly available, with licenses that allow academic use.
We used the datasets for exclusively academic purposes.

\subsection{Dysarthric Speech Datasets}\label{subsubsec:dysarthric}
\textbf{UASpeech} \citep{kim2008dysarthric} comprises recordings from 25 English speakers, of whom 14 have dysarthria and 11 are healthy. The severity of dysarthria was assessed using the Frenchay Dysarthria Assessment (FDA) \citep{enderby1980frenchay}, categorizing four speakers as having high-intelligibility, three as mid-intelligibility, three as low-intelligibility, and four as very low-intelligibility. This study analyzes common and uncommon words selected for their diverse phonetic sequences, which are crucial for evaluating the pronunciation of phonemes in varied contexts. Although the dataset includes recordings from an 8-microphone array, we utilize only the fifth microphone for computational efficiency. In total, 6,589 utterances from healthy speakers and 8,370 utterances from dysarthric speakers are used. 
For time alignment information, we apply the Montreal Forced Aligner (MFA) \citep{mcauliffe2017montreal}.

\textbf{TORGO} \citep{rudzicz2012torgo} consists of recordings from 15 English speakers, including 8 with dysarthria and 7 healthy individuals. The severity was determined using FDA scores, classifying two speakers as mild, one as mild-to-moderate, one as moderate-to-severe, and four as severe. To balance the classes, mild-to-moderate and moderate-to-severe speakers were merged into a single moderate category. A total of 156 healthy and 413 dysarthric utterances are used.
Similar to UASpeech, we apply MFA for time alignment information. 
Then, we integrate alignments that were manually adjusted by two linguists, as outlined by \citet{hernandez2020dysarthria}. 
These manually refined alignments will be made publicly accessible in our repository.$^1$

\textbf{SSNCE} \citep{ta2016dysarthric} comprises Tamil speech recordings from 20 dysarthric and 10 healthy speakers. Severity is categorized based on intelligibility scores rated on a 7-point Likert scale by two speech pathologists, resulting in 7 mild (scores 1-2), 10 moderate (scores 3-4), and 3 severe (scores 5-6) speakers. Each speaker recorded 260 distinct sentences, totaling 2,600 healthy and 5,200 dysarthric utterances. 
We use the time stamps provided with the datasets.

\subsection{Nonnative Speech Datasets}\label{subsubsec:nonnative}
\textbf{speechocean762} \citep{speechocean762} consists of 5000 utterances from 250 Mandarin-speaking non-native children and adult speakers. 
English proficiency levels' ratio maintains a 2:1:1 ratio for good, medium, and poor, ensuring representation across different proficiencies.
Our analysis focuses on total scores at the sentence level, pronunciation quality graded on a scale from 0 to 10.
We use scores 9 and 10 as training data and the rest as test data.
We employ forced alignments generated using the Kaldi recipe, following the experimental setup of \citet{speechocean762}. 
This allows for a direct comparison of S3M and Kaldi features, as the quality of phoneme alignment can affect the evaluation results \citep{mackenzie2020assessing}.

\textbf{L2-ARCTIC} \citep{zhao2018l2} is a non-native English corpus comprising recordings from 24 speakers, whose first languages (L1) include Hindi, Korean, Mandarin, Spanish, Arabic, and Vietnamese. 
Each speaker contributes approximately one hour of read speech derived from CMU ARCTIC prompts \citep{kominek2004cmu}.
For this study, we use 150 utterances per speaker, where manual annotations were available for phonememic errors such as substitutions, deletions, and additions.
We use the existing data split as the train/test split, where we further exclude mispronounced utterances in the training data.
Unlike other datasets, L2-ARCTIC only contains phoneme-wise mispronunciation detection labels (0/1).
Therefore, we used the GoP scores and the label at the phoneme level, not at the utterance level.
We use the time stamps provided with the datasets.

\section{Computational cost}
Extracting S3M features for all the datasets takes less than one day for each S3M on a single NVIDIA V100 GPU.
Running the experiments on all the datasets takes less than a day on a 64-CPU 128GB-memory machine for the default hyperparameter settings.

\section{Additional analyses and discussions}\label{sec:add-analysis}

\subsection{Layerwise analysis of downstream performance}\label{subsec:layerwise}
\begin{figure*}[t]
    \centering
    \subfloat[UASpeech]{\includegraphics[height=0.3\textwidth]{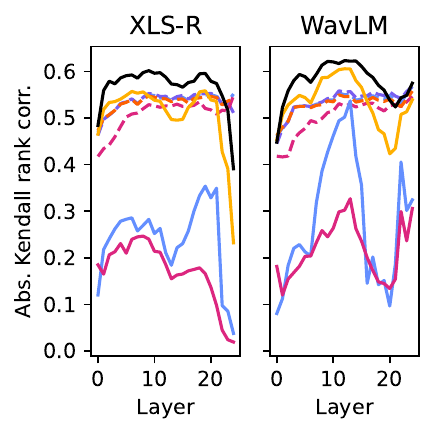} \label{fig:layerwise_uaspeech}}
    \subfloat[TORGO]{\includegraphics[height=0.3\textwidth]{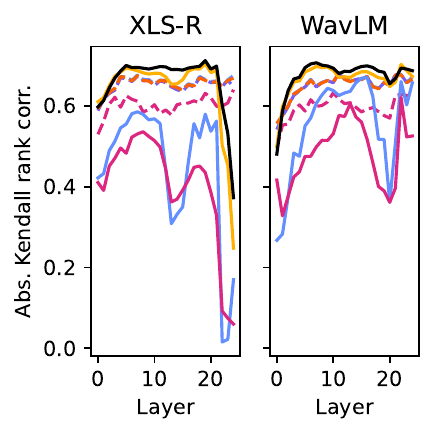} \label{fig:layerwise_torgo}}
    \subfloat[SSNCE]{\includegraphics[height=0.3\textwidth]{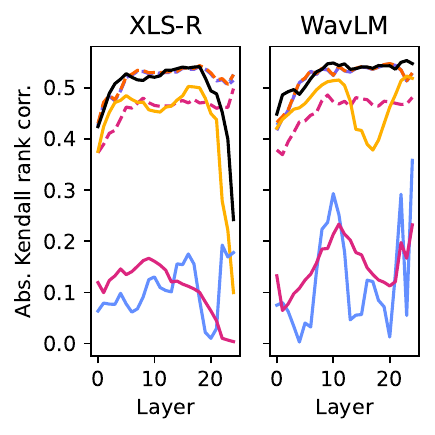} \label{fig:layerwise_ssnce}}\\
    \subfloat[speechocean762]{\includegraphics[height=0.3\textwidth]{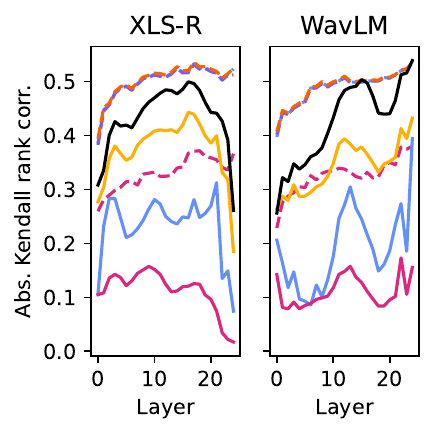} \label{fig:layerwise_so762}}
    \subfloat[L2-ARCTIC]{\includegraphics[height=0.3\textwidth]{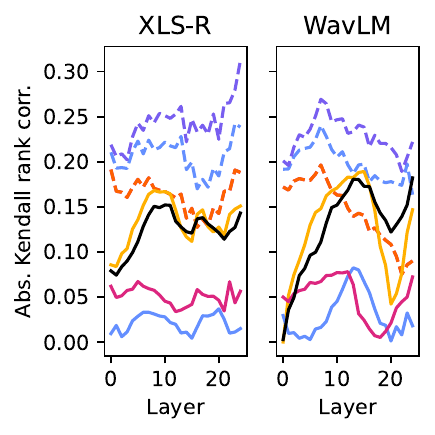} \label{fig:layerwise_l2arctic}}\\
    \vspace{-1em}\subfloat{\includegraphics[width=0.9\textwidth]{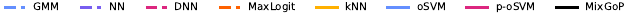}}
    \caption{
Kendall-tau correlation coefficient when features are extracted from different layers of S3M models.
    }
    \label{fig:layerwise}
\end{figure*}
It is known that different layers of S3Ms tend to encode different information \citep{pasad2021layer,pasad2023self,pasad2023comparative,choi2024self}.
Hence, we further compare the layerwise trend of XLS-R and WavLM in \Cref{fig:layerwise} to provide a guideline for which layer to use for each downstream task.
WavLM tends to have decent or even the best performance in the final layer, while XLS-R features often suffer a great decrease in performance in the final layer.
This indicates that the choice of which layer to use is more critical when using XLS-R compared to WavLM, whereas the last layer of WavLM generally shows decent performance.

\subsection{Impact of the number of subclusters}\label{subsec:abl-clusters}
\begin{table}[]
\caption{Ablation on the number of clusters for Gaussian mixtures.}
\label{tab:ncluster}
\vspace{0.5em}
\resizebox{1.0\linewidth}{!}{%
\begin{tabular}{l|ccccc}
\toprule
Dataset & 4 & 8 & 16 & 32 & 64 \\
\midrule
UASpeech & 0.613 & 0.620 & 0.621 & \textbf{ 0.623 } & 0.622 \\
TORGO & 0.704 & 0.703 & 0.706 & \textbf{ 0.713 } & N/A \\
SSNCE & 0.544 & 0.548 & \textbf{ 0.553 } & \textbf{ 0.553 } & N/A \\
speechocean762 & \textbf{ 0.545 } & 0.543 & 0.544 & 0.539 & 0.537 \\
L2-ARCTIC & 0.175 & 0.178 & 0.176 & \textbf{ 0.182 } & 0.179 \\
\bottomrule
\end{tabular}
}
\end{table}
We explored the optimal number of subclusters for the downstream task.
We conducted a grid search with cluster sizes of 4, 8, 16, 32, and 64, while keeping the best model and layer index from \Cref{subsec:results} fixed.
For the TORGO and SSNCE datasets, we did not test 64 clusters due to insufficient phoneme samples to train the Gaussian mixtures.

As shown in \Cref{tab:ncluster}, increasing the number of clusters often results in marginal improvements in downstream performance. 
We attribute this to the fact that better distribution modeling tends to enhance performance.
This reinforces our hypothesis that the number of subclusters does not need to exactly match the number of allophones, as sufficiently large number of Gaussian mixtures can approximate any probability density \citep{nguyen2020approx}.
The best performance was consistently achieved with a cluster size of 32 across datasets, with the exception of the speechocean762 dataset. 
For speechocean762, the highest performance was attained with the smallest cluster size, 4, although performances with cluster sizes of 8 and 16 were also similar.

\begin{figure}[t!]
    \centering
    \includegraphics[width=0.45\textwidth]{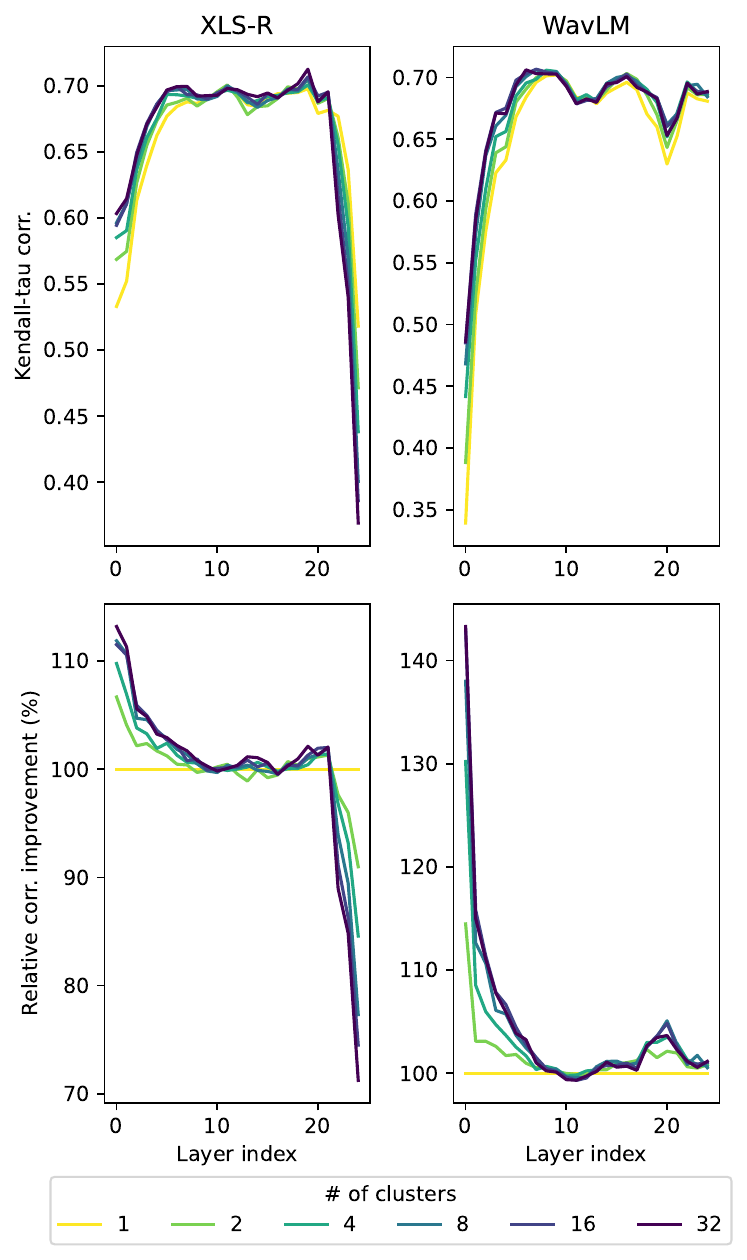}
    \caption{
        Downstream performances with varying number of clusters and S3M layer index.
        Relative performance improvement is obtained by dividing the Kendall-tau correlation with that of cluster size 1.
        Except for the later layers of XLS-R with extreme performance degradation, having bigger number of clusters yield better performances.
    }
    \label{fig:cluster_abl}
\end{figure}

To further analyze the behaviors of different clusters, we performed additional layerwise analysis on TORGO, the smallest dataset, in \Cref{fig:cluster_abl}.
We can clearly observe that a larger number of clusters leads to better downstream performance across different layers.
Especially on the best-performing layer (layer index 19 of XLS-R and 6 for WavLM), it shows bigger differences per different number of clusters.
We can also observe that the number of clusters 16 and 32 is the most similar, potentially indicating the performance saturation with respect to the number of clusters.


\subsection{Learnable phoneme-wise attention}\label{subsec:attn}
In dysarthric speech, pronunciation scores for certain phonemes are known to exert more influence on speech intelligibility \citep{yeo23_interspeech}. 
Relevant factors include their place and manner of articulation, and articulatory complexity \citep{kim2010frequency}.
Non-native speakers, on the other hand, make different pronunciation mistakes based on their native language \citep{ng2023l1,yeo2023comparison}.

To model the phoneme-wise importance, we design a learnable attention module $\alpha \in \mathbb{R}^{|\mathcal{V}|}$ to satisfy two conditions: (i) bounded attention weights ($0 \leq \alpha[p] \leq 1$ for any phoneme $p$), and (ii) the weights sum up to one ($\sum_{i=1}^{N} \alpha[p_i] = 1$):
\begin{align}
    \alpha[p_i] = \frac{e^{\mathbf{w}_{p_i}}}{\sum_{i=1}^N e^{\mathbf{w}_{p_i}}},\label{eq:attn}
\end{align}
where the phoneme-wise logits $\mathbf{w} \in \mathbb{R}^{|\mathcal{V}|}$ is a learnable parameter with the vocabulary size.
The formulation is nearly the same as the \texttt{softmax} function with the minor difference: if the same phoneme occurs multiple times within the utterance, it shares the same weights.

We can extend our MixGoP by combining the Gaussian mixtures (\Cref{eq:GM}) and the attention module (\Cref{eq:attn}):
\begin{align}
    \texttt{MixGoPAttn}(\mathbf{x}) := \sum_{i=1}^{N} \alpha[p_i] \cdot \log P_\theta(\mathbf{s}|p).\label{eq:mixtures}
\end{align}
We improve upon the phoneme-wise GoP of \Cref{eq:1_n} by (i) replacing the phoneme classifier $P_\theta(p|\mathbf{s})$ by the phoneme density estimator $P_\theta(\mathbf{s}|p)$ and (ii) replacing the uniform importance $1/N$ with the learnable weight $\alpha[p_i]$.

We train the phoneme-wise attention module by directly maximizing the Spearman's rank correlation coefficient 
between $\texttt{MixGoPAttn}(\mathbf{x})$ and the pronunciation score $y$ (degree of dysfluency/disfluency for $\mathbf{x}$):
\begin{align}
    \mathcal{L} = - \texttt{spearman}(\text{MixGoPAttn}(\mathbf{x}), y),
\end{align}
where we freeze the Gaussian mixture models and only train the logits $\mathbf{w}$.
The sorting operation within \texttt{spearman} is made differentiable by soft sorting \citep{blondel2020fast}, following \citet{blondel2020fast}'s implementation. 

\begin{figure}[t]
    \centering
    \includegraphics[width=0.43\textwidth]{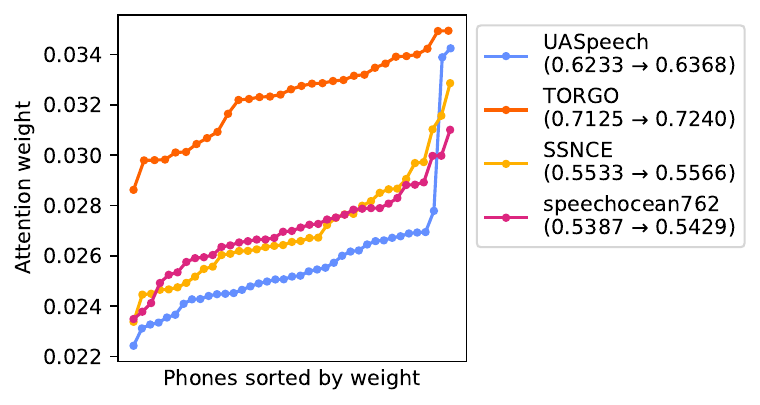} 
    \caption{
    Learned attention scores and the performance improvement on UASpeech, TORGO, SSNCE, and speechocean762.
    }
    \label{fig:attn}
\end{figure}

   

We applied 5-fold cross-validation to obtain accurate evaluation results.
The phoneme attention module is trained on four datasets, excluding L2-ARCTIC as the dataset does not provide utterance-level scores.
Note that this setting differs from the experiments demonstrated in \Cref{tab:main}, as we use the test set to train the attention module.

As demonstrated in \Cref{fig:attn}, applying the attention module resulted in small yet consistent performance improvements across all datasets. 
Furthermore, the attention weight differences between the least and most influential phonemes varied by up to 1.5 times, indicating variability in their contributions. 
This highlights the importance of considering the differing influence of each phoneme when calculating utterance-level pronunciation scores, which is crucial for optimizing pronunciation assessment performance.

However, we could not identify a consistent pattern across datasets regarding which phonemes consistently received higher or lower attention scores. 
This suggests that the variation in attention weights may be influenced by various factors, such as speakers, and phoneme distribution.
Further investigation into the underlying mechanisms driving these variations is necessary to gain a deeper understanding of the differing impact of phonemes in pronunciation assessment.\footnote{Full list of attention weights can be found in \url{https://github.com/juice500ml/acoustic-units-for-ood}}

\subsection{Comparing MixGoP and kNN}\label{subsec:gop_vs_knn}
Both kNN and our proposed MixGoP heavily depend on the distances induced by S3M features.
Also, kNN is often the closest competitor to MixGoP, as shown in \Cref{tab:main}, indicating their similarities.
However, there are multiple differences in their implementation details.
kNN relies on Euclidean distance and selects the maximum distance among nearest neighbors.
MixGoP employs Mahalanobis distance and measures the distances from the centroids obtained by the EM algorithm.
We leave the study on the effectiveness of kNN and MixGoP's key components for future work.





\end{document}